\RequirePackage{fix-cm}
\documentclass[smallextended]{svjour3}       
\smartqed  
\usepackage{graphicx}
\usepackage{multirow}
\usepackage{float}
\usepackage{subfigure}
\usepackage{tabularx}
\usepackage{makecell}
\usepackage{tabulary}
\usepackage{chngpage}
\usepackage{fancyhdr}
\usepackage{multirow}
\usepackage{amsfonts}
\usepackage{microtype}
\usepackage{array}
\newcolumntype{?}[1]{!{\vrule width #1}}
\usepackage{url}

\usepackage{tikz}

\begin{document}

\title{Face Detection in the Operating Room: Comparison of State-of-the-art Methods and a Self-supervised Approach  
}

\titlerunning{Face Detection in the Operating Room}        

\author{Thibaut Issenhuth         \and
        Vinkle Srivastav \and  Afshin Gangi \and Nicolas Padoy 
}



\institute{T. Issenhuth \at
              \email{issenhuth@unistra.fr}           
           \and
}

\institute{Thibaut Issenhuth \and Vinkle Srivastav \and Nicolas Padoy \at
              ICube, University of Strasbourg, CNRS, IHU Strasbourg, France \\
              \email{issenhuth@unistra.fr . srivastav@unistra.fr . npadoy@unistra.fr}          
         \and 
         Afshin Gangi \at Radiology Department, University Hospital of Strasbourg, France
}

\date{Received: date / Accepted: date}

\maketitle
\begin{abstract}
\emph{Purpose:} Face detection is a needed component for the automatic analysis and assistance of human activities during surgical procedures. Efficient face detection algorithms can indeed help to detect and identify the persons present in the room, and also be used to automatically anonymize the data. However, current algorithms trained on natural images do not generalize well to the operating room (OR) images. In this work, we provide a comparison of state-of-the-art face detectors on OR data and also present an approach to train a face detector for the OR by exploiting non-annotated OR images.\\
\emph{Methods:} We propose a comparison of 6 state-of-the-art face detectors on clinical data using Multi-View Operating Room Faces (MVOR-Faces), a dataset of operating room images capturing real surgical activities. We then propose to use self-supervision, a domain adaptation method, for the task of face detection in the OR. The approach makes use of non-annotated images to fine-tune a state-of-the-art detector for the OR without using any human supervision.\\
\emph{Results:} 
The results show that the best model, namely the tiny face detector, yields an average precision of 0.556 at Intersection over Union (IoU) of 0.5. Our self-supervised model using non-annotated clinical data outperforms this result by 9.2\%.\\
\emph{Conclusion:} We present the first comparison of state-of-the-art face detectors on operating room images and show that results can be significantly improved by using self-supervision on non-annotated data.\\

\end{abstract}

\section{Introduction}
\begin{figure}
\centering
\subfigure{\includegraphics[width=0.32\linewidth]{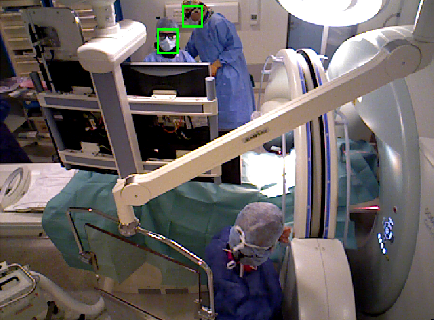}}
\subfigure{\includegraphics[width=0.32\linewidth]{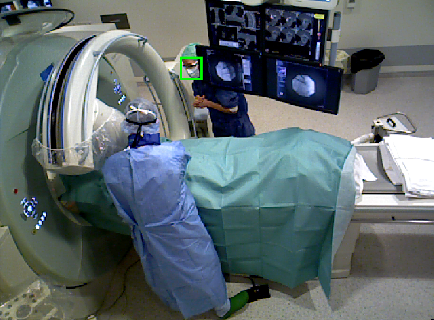}}
\subfigure{\includegraphics[width=0.32\linewidth]{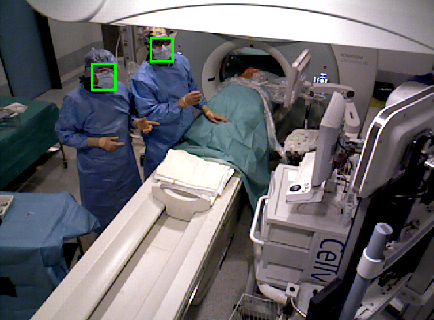}}
\\[-0.8ex]
\subfigure{\includegraphics[width=0.32\linewidth]{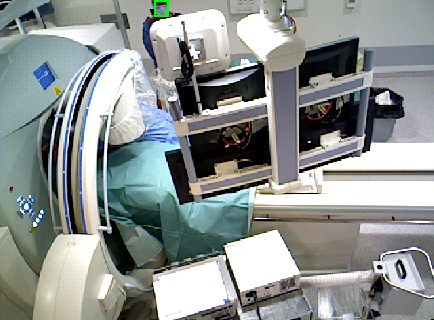}}
\subfigure{\includegraphics[width=0.32\linewidth]{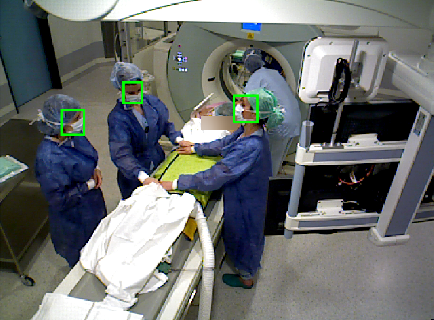}}
\subfigure{\includegraphics[width=0.32\linewidth]{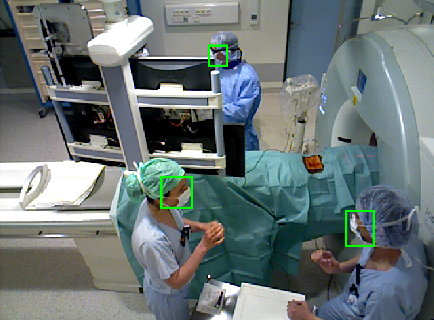}} \\[-2ex]
\caption{Examples of images collected in the OR, illustrating the challenges for face detection systems: occlusion with medical equipment or other persons, masked faces, absence of visible skin. Ground-truth is shown with green bounding boxes.}
\label{fig:mvor-faces-challenges}
\end{figure}
\indent Modern ORs are high-tech environments, where surgical activities are increasingly captured with cameras. The automatic understanding of this visual data by extracting rich and meaningful information is a promising way of developing machine intelligence and context-aware systems in the clinical environment. It will help to improve the workflow in hospitals by developing better decision support systems for the clinicians \cite{chen2018patient}. Face detection in the operating room is one of the key steps needed to develop intelligent context-aware systems for the automatic analysis of human activities. It can indeed serve for person detection and identification as well as for the anonymization of sensitive OR data. \\
\indent Face detection is a very active research topic in computer vision. Before the rise of deep learning, traditional methods for face detection used machine learning algorithms on top of hand-crafted features \cite{viola2001rapid}. With the advent of deep learning architectures based on convolutional neural networks (CNNs), the performance of face detectors has drastically improved. CNNs are trained end-to-end and are able to learn semantically rich and robust data representations that yield great accuracy. The face detection architectures are often inspired by deep object detectors, whether they are one-stage \cite{najibi2017ssh,1708.05237} or two-stage detectors \cite{1606.03473}. The one-stage detectors generally divide the image into a grid of boxes and directly classify and regress the localization of objects in each box. The two-stage networks first use a Region Proposal Network (RPN) \cite{ren2015faster} to extract Region of Interests (ROIs), then a second network to classify and localize each ROI more accurately. These detectors manage to handle the variety of scales, by setting up strategies to specifically detect small faces. They perform contextual reasoning and use image or features pyramids to achieve robustness. The success of these methods can also be attributed to the availability of large-scale annotated dataset. Indeed, WIDER Faces \cite{yang2016wider}, the standard dataset for training and testing face detection methods in the wild, contains 32,203 images with 393,703 labeled faces. Apart from the bounding box based face detection methods, faces can also be extracted from the face keypoints of human pose estimators, which is performed by mainly two types of approaches: bottom-up and top-down. Bottom-up approaches \cite{cao2016realtime,insafutdinov2016deepercut} first detect all the keypoints, then assemble them into skeletons, whereas top-down approaches \cite{fang2017rmpe,xiao2018simple,Chen2018CPN} first detect persons, often with standard object detectors, and then detect keypoints for each detected person using a single person pose estimator. The top-down approaches resolve better the keypoint to person assignment and therefore largely outperform bottom-up approaches on the standard public datasets \cite{lin2014microsoft,andriluka14cvpr}. \\
\indent The automatic recognition of activities during real surgeries to develop intelligent context-aware assistance systems is a recent field that has started to gain traction in the medical as well as computer vision community \cite{twinanda2016m2cai,maier2017surgical,yeung2018bedside}. Work on analyzing humans in OR videos have generally focused on person bounding box detection and on human pose estimation, using either RGB or RGB-D data \cite{kadkhodamohammadi2017-ar,Kadkhodamohammadi2017-tx,belagiannis2016parsing}. So far, face detection in the OR has however received very little attention, besides the recent work \cite{1808.04440}, described below.  
Furthermore, current state-of-the-art face detectors, even those close to the human-level performance, do not generalize well to OR images. Their inferior generalization can be explained by the fact that they have been trained on natural images, whereas OR images are very specific and challenging: persons' faces are often occluded due to equipment clutter, masks, and glasses. Figure \ref{fig:mvor-faces-challenges} shows some examples of the challenging situations occurring inside the OR.

 One standard approach to overcome this visual domain difference is to use transfer learning, which adapts a method by fine-tuning its parameters on an annotated target dataset. In \cite{1808.04440}, authors recently proposed such a method to detect the faces in the OR by finetuning the Faster-RCNN detector \cite{1606.03473} on OR videos. The video dataset consists of youtube OR videos, which have been manually annotated with face bounding boxes. In  \cite{1808.04440}, the results are further improved by using temporal smoothing. Manual annotation can, however, be expensive and time-consuming, whereas non-annotated data is in abundance and often inexpensive. Therefore, this work aims at distilling knowledge from non-annotated OR data to improve the baseline performance of a face detector.\\
\indent This paper investigates face detection and visual domain adaptation for the OR environment. We first present a comparison of 6 state-of-the-art face detectors. We consider methods where faces can be obtained either directly from bounding box based methods or from face keypoints generated by human pose estimators. We evaluated these methods on the MVOR-Faces
, an extension of the MVOR dataset \cite{srivastav2018mvor} augmented with face bounding box annotations.
To the best of our knowledge, this paper presents the first comparison of state-of-the-art face detectors in an OR environment. We also select one detector, SSH \cite{najibi2017ssh}, and propose to improve it by using an iterative self-supervised method. Several variants of self-supervised methods have been recently used to improve the quality of the synthetic annotations, for instance by using temporal ensembles \cite{laine2016temporal} or by combining the results of different geometric transformations \cite{1712.04440}. In this work, we found it effective to iteratively generate synthetic annotations and fine-tune the model. This approach significantly improves the original model, and largely outperforms the best face detectors on all metrics.

\section{Methodology}
\begin{figure}
    \centering
    \includegraphics[scale=0.34]{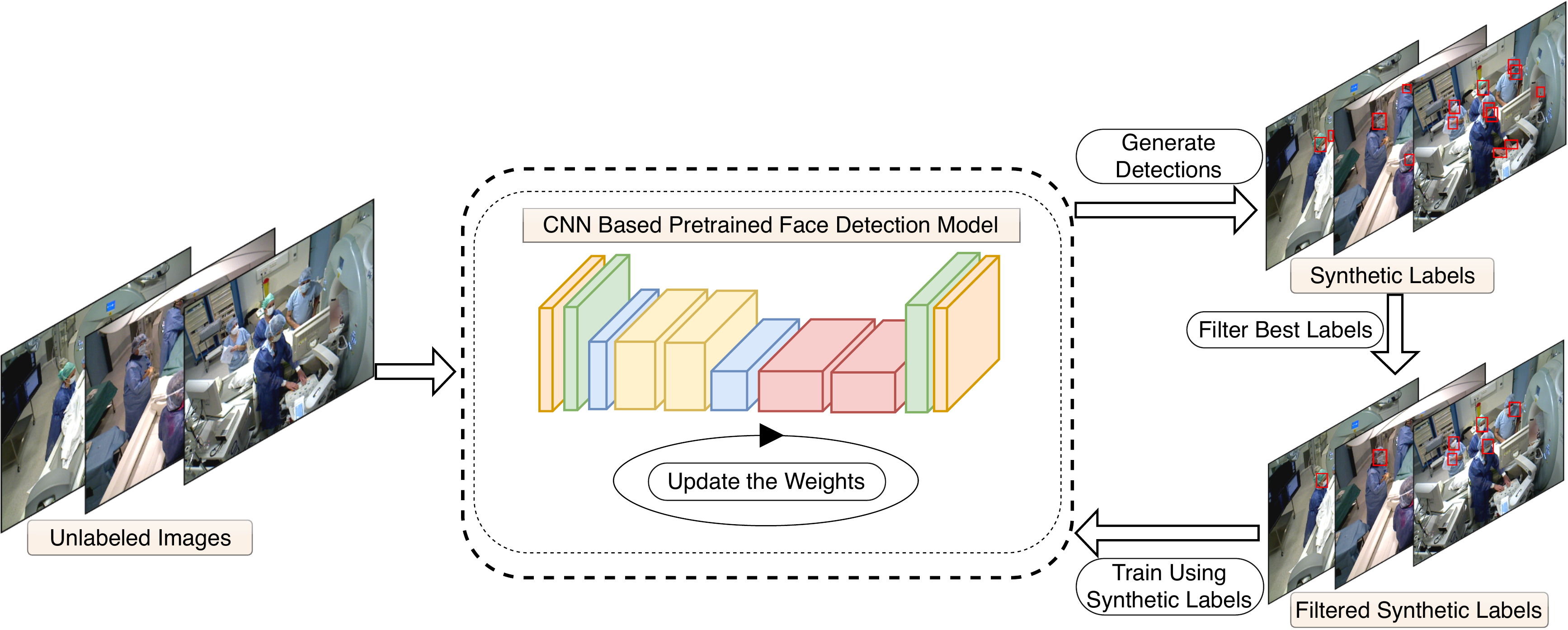}
    \caption{An iterative approach to adapt a face detector to a target operating room. First, we obtain a trained face detector (SSH \cite{najibi2017ssh}) and unlabeled images of the same operating room. Then, we repeat the following steps: (a) use the detector to generate the labels (b) filter the detections with a heuristic to create good quality synthetic annotations (c) retrain the detector using synthetically generated annotations.}
    \label{fig:iterative_approach}
\end{figure}
\subsection{\textbf{Comparison of State-of-the-art Face Detector}}
We present below the state-of-the-art methods for face detection used in our comparison. In this study, the faces are represented by bounding boxes. We consider 4 methods where the faces are directly obtained as the output of the detector and 2 methods where the face bounding boxes are generated from face keypoints detected by human pose estimators. These methods are selected based on their ranking on standard public datasets, namely the WIDER dataset \cite{yang2016wider} for bounding box based face detectors and the COCO dataset \cite{lin2014microsoft} for human pose estimators. For reproducibility, we only choose open-source methods. 
\subsubsection{\textbf{Bounding Box Based Face Detectors}}
\textbf{Faster-RCNN face detector \cite{1606.03473}}. The Faster-RCNN, originally designed as a generic two-stage object detector, was trained for the face detection task on the WIDER Faces dataset. First, the RPN generates ROIs with a sliding window approach on deep feature maps. At each sliding window location, anchors, bounding boxes of different scales and aspect ratios, are predicted as either background or ROI. Then, ROIs are pooled and used as input for a second network, which classifies the face and regresses for the exact coordinates of its bounding box.\\
\textbf{Finding tiny faces \cite{Hu_2017_CVPR}}. This method is specifically conceived to detect faces of different scales. The input of the algorithm is an image pyramid, with three versions of the image: one downsampled, the original and one upsampled image. Each rescaled image is processed by a shared pyramidal CNN, which predicts binary heatmaps for bounding box templates of different sizes. \\
\textbf{SSH: Single Stage Headless Face Detector \cite{najibi2017ssh}}. This is a one-stage face detector that includes a context module, namely a set of convolutional layers to increase the effective receptive field and different branches to achieve scale-invariance. It uses three different detector networks to predict small, medium and large face anchors. SSH achieves a similar accuracy than the tiny face detector \cite{Hu_2017_CVPR}, while maintaining real-time performance.\\ 
\textbf{S$^3$FD: Single Shot Scale-invariant Face Detector \cite{1708.05237}}: This is also a one-stage face detector inspired by the RPN \cite{ren2015faster} and SSD \cite{liu2016ssd} architectures. They design strategies to increase the number of positive anchors matching tiny faces during training. Their CNN architecture includes feature maps and detectors that are specific to a range of scales and uses a max-out background label on small anchors to reduce the number of false positives.
\\[-4ex]
\subsubsection{\textbf{Human Pose Estimation Based Face Detectors}}
Human pose estimation aims to localize the anatomical keypoints of all the persons present in an image. These anatomical keypoints are spread across the whole body including the face keypoints. Current state-of-the-art methods are trained on the COCO dataset, which includes five face keypoints (nose, left eye, right eye, left ear, right ear). We fit a bounding box of fixed size, 30x30 pixels, at the average location of these five keypoints to extract the face bounding box. We now describe the pose estimation methods that we used for the evaluation.\\
\textbf{OpenPose \cite{cao2016realtime}}. This is one of the best bottom-up approaches for human pose estimation. First, a CNN predicts confidence heat-maps for all keypoints and part affinity fields for each joint. These maps are then used to construct a graph of keypoints and body joints. Finally, this graph is parsed using a bi-partite graph matching algorithm to produce a set of human poses. \\
\textbf{AlphaPose \cite{fang2017rmpe}}. This is one of the state-of-the-art top-down methods. It uses Faster-RCNN \cite{ren2015faster} to detect persons. Then, cropped human bounding boxes are processed by a single person pose estimator, which is composed of several modules, including spatial transformers, to refine the keypoint detections in the bounding box. This method successfully handles the problems caused by inaccurate and duplicate bounding boxes.

\subsection{\textbf{Iterative Self-supervised Approach for Face Detection in the OR}}
We use the following two steps to improve the selected state-of-the-art face detector on OR images: 1. Generation of the unlabeled dataset 2. Iterative refinement using a self-supervised approach. \\
\subsubsection{\textbf{Generation of the Unlabeled Dataset}}
We use an unlabeled dataset of 20k images generated from videos captured in the OR. The videos were collected on days different from the ones of the test dataset to ensure the absence of overlap between the unlabeled dataset and the test dataset. We then use OpenPose \cite{cao2016realtime}, a multi-person pose estimator, on the OR videos to get the approximate number of persons in each frame. The computational efficiency of OpenPose allows us to make the inference on the entire dataset in a reasonable time. We divide the images into four categories: images with one, two, three, and four or more detected persons. Since OpenPose also gives a confidence score for each detected skeleton, we average the scores of the detected skeletons and take the 5k highest-scored images from each category (i.e., 20k images overall). This selection method ensures that the images contain persons in different numbers.

\subsubsection{\textbf{Iterative Refinement using Self-supervised Approach}} 
We utilize an iterative self-supervised approach to adapt the state-of-the-art model to the target OR dataset. This approach consists of fine-tuning the model on a subset of its own detections. We use SSH \cite{najibi2017ssh}, pre-trained on WIDER Faces, as the CNN-based model for the iterative refinement. We choose this model because it has high computational efficiency and also yields state-of-the-art results on WIDER Faces. This detector is then used to generate synthetic labels on the unlabeled dataset. To select quality face bounding boxes, we use a simple yet effective heuristic criteria: with a dataset of N images, we select the best 2*N detections. Since we have approximately 2.5 persons/image in the unlabeled dataset, 2*N best detections contain the face bounding boxes with a high recall. These synthetically annotated images are then used to finetune the original model. We perform these steps iteratively to improve the detections and the detector at each iteration as shown in Fig. \ref{fig:iterative_approach}. It is to be noted that no validation set is available as we did not use any supervised annotation. Therefore, our experiments differ from the traditional deep learning experiments, which fine-tune the hyper-parameters based on the performance on a validation set. We mainly conduct the fine-tuning experiments with a different number of training batches before relabelling and different iteration numbers. We present the result of each experiment on the test-set.
\section{Experimental Setup}
\subsection{Test Dataset (MVOR-Faces)}
We compare the state-of-the-art face detectors on MVOR-Faces, a dataset of operating room images captured during real surgical procedures. MVOR-Faces is an extension of the public MVOR dataset \cite{srivastav2018mvor}, which consists of 732 multi-view frames (2196 images) recorded in an interventional room. In the MVOR dataset, faces of persons without a mask and nude parts of patients are fully blurred, and the persons with masks are blurred only on the eyes. MVOR-Faces contains the same images as MVOR, except that the eyes of the persons wearing a mask are not blurred. Also, it contains the manually annotated face bounding box of all visible faces wearing a mask. All fully-visible faces and nudity zones are still blurred in the MVOR-Faces as needed for anonymity. Overall, the dataset contains 2262 face bounding boxes for 2196 images.
\subsection{Evaluation Metrics} 
We use the standard metrics for object detection from COCO \cite{lin2014microsoft}, i.e. Average Precision (AP) and Average Recall (AR). $AP^{IoU}$ is the average precision at a fixed intersection over union (IoU), and  AP is the average of $AP^{IoU}$ at different IoU thresholds. While the public implementation averages between IoU of 0.5 and 0.95 with a step of 0.05, we average between an IoU of 0.3 and 0.95 to support a slightly looser metric. The consideration of a slightly looser metric is motivated by the fact that face detection in a medical context is quite challenging: clinicians wear mask, glasses, and hats, and are often occluded. Therefore, a looser metric reduces the bias in favor of the face detectors.
\begin{figure}
\centering
\subfigure[Faster-RCNN \cite{ren2015faster}]{\includegraphics[width=0.29\linewidth]{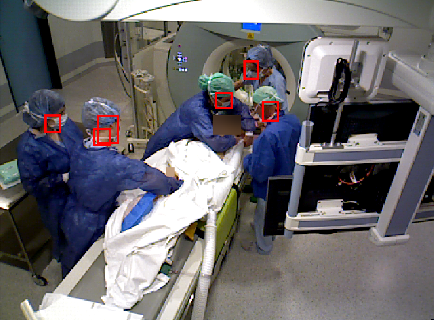} \includegraphics[width=0.29\linewidth]{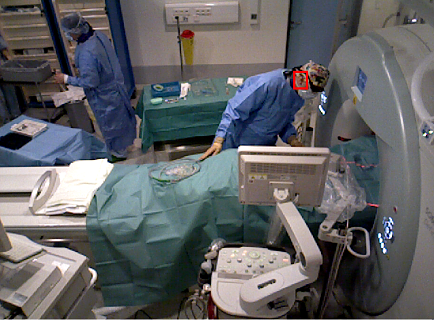}}
\subfigure[S3FD \cite{1708.05237}]{\includegraphics[width=0.29\linewidth]{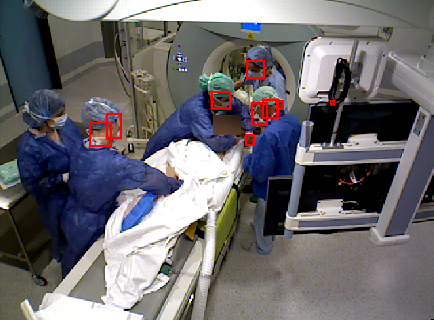} \includegraphics[width=0.29\linewidth]{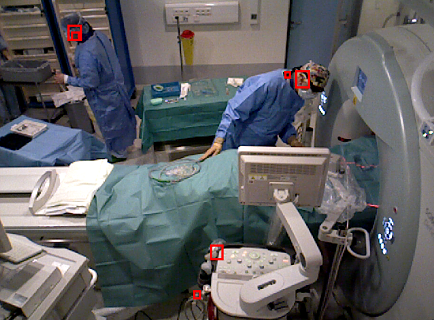}}
\subfigure[Tiny Face \cite{Hu_2017_CVPR}]{\includegraphics[width=0.29\linewidth]{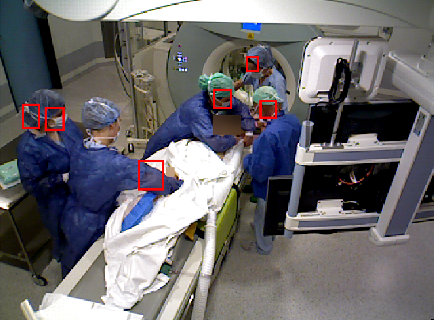} \includegraphics[width=0.29\linewidth]{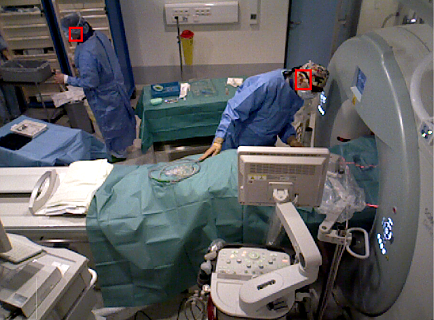}}
\subfigure[SSH \cite{najibi2017ssh}]{\includegraphics[width=0.29\linewidth]{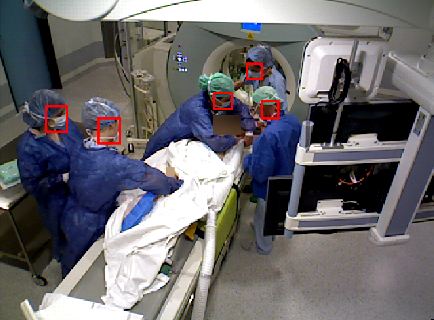} \includegraphics[width=0.29\linewidth]{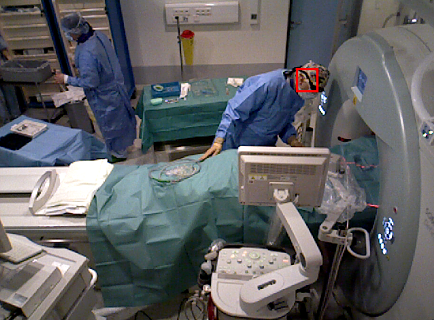}}
\subfigure[AlphaPose \cite{fang2017rmpe}]{\includegraphics[width=0.29\linewidth]{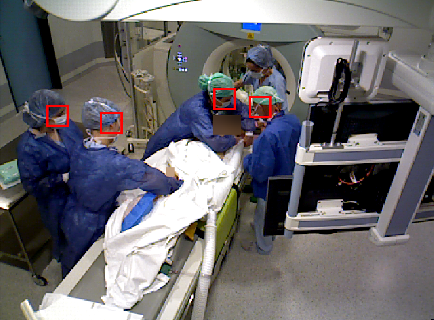} \includegraphics[width=0.29\linewidth]{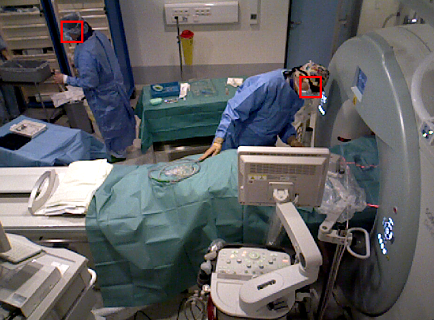}}
\subfigure[OpenPose \cite{cao2016realtime}]{\includegraphics[width=0.29\linewidth]{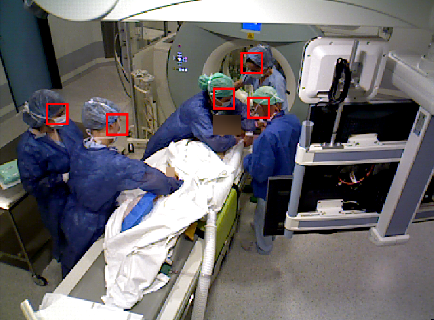} \includegraphics[width=0.29\linewidth]{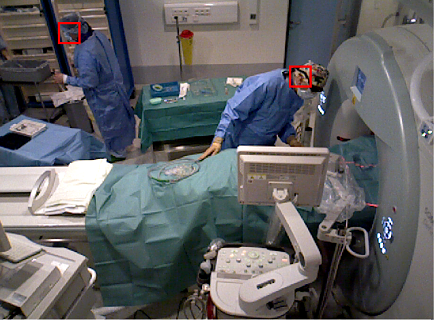}}
\caption{Qualitative results from the face detectors evaluated on MVOR-Faces. The displayed detections were selected based on a score threshold of the detector corresponding to a recall threshold of 70\% at an IoU of 0.3. }
\label{qualitative_soa}
\end{figure}

\begin{figure}
\centering
\subfigure{\includegraphics[width=0.39\linewidth]{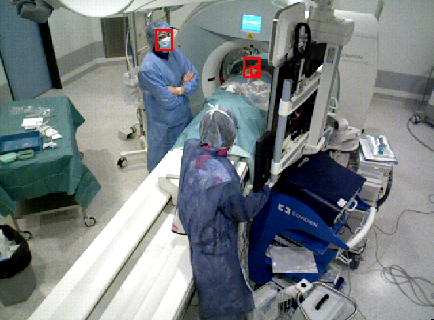}}
\subfigure{\includegraphics[width=0.39\linewidth]{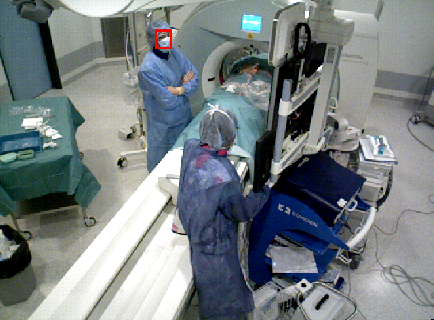}}
\\[-0.9ex]
\subfigure{\includegraphics[width=0.39\linewidth]{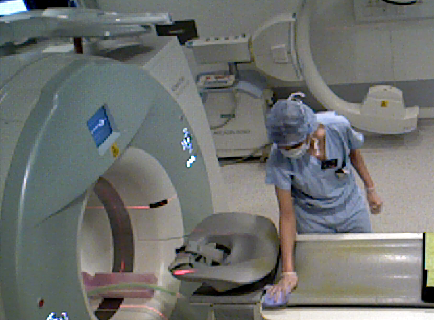}}
\subfigure{\includegraphics[width=0.39\linewidth]{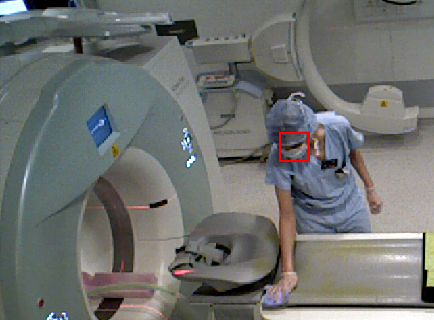}}
\\[-0.9ex]
\subfigure{\includegraphics[width=0.39\linewidth]{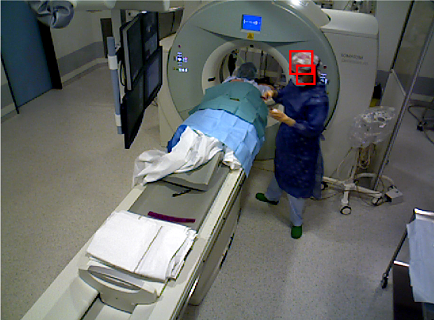}}
\subfigure{\includegraphics[width=0.39\linewidth]{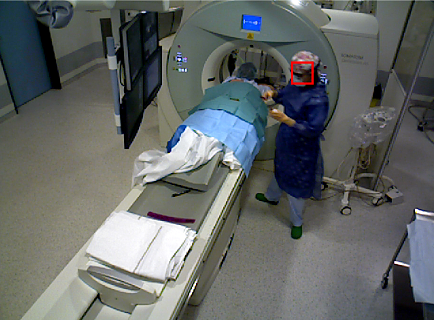}}
\\[-0.9ex]
\subfigure{\includegraphics[width=0.39\linewidth]{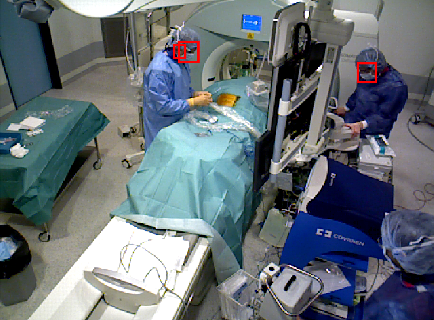}}
\subfigure{\includegraphics[width=0.39\linewidth]{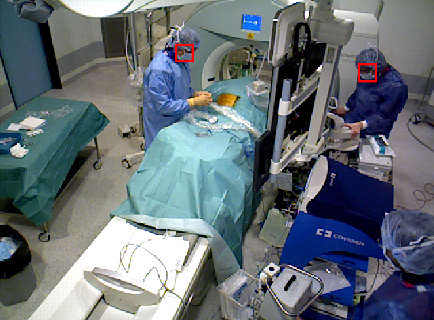}}
\\[-0.9ex]
\subfigure{\includegraphics[width=0.39\linewidth]{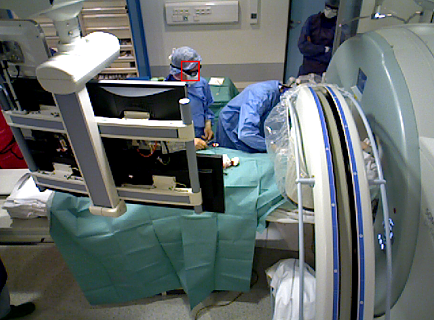}}
\subfigure{\includegraphics[width=0.39\linewidth]{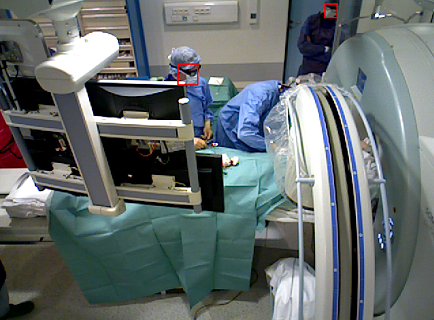}}
\caption{Comparison of the original SSH model (left column) with the best self-supervised model trained with our iterative approach (right column). To filter the displayed detections, we use the score threshold corresponding to a recall threshold of 70\% at an IoU of 0.3, as in Fig. \ref{qualitative_soa}. The self-supervised model detects much harder examples, with occlusion or uncommon poses. }
\label{Qualitative results}
\end{figure}

\section{Results and Discussion}
\subsection{Comparison of State-of-the-art Face Detectors}
Table \ref{state_of_the_art} shows the comparison of state-of-the-art face detectors. The first four methods in Table \ref{state_of_the_art} directly output face bounding boxes, as described in section 2.1.1. The next two methods detect the human skeletons, including face keypoints (i.e. ears, eyes and nose). We extract the face bounding boxes from face keypoints as specified in Section 2.1.2. Unless otherwise stated, we use the exact same models provided by the authors, without modifying any hyper-parameter. \\
On AP(0.3:0.95), Tiny Face detector \cite{Hu_2017_CVPR} is the best model, with 0.340; SSH \cite{najibi2017ssh} and S3FD \cite{1708.05237} are close, with respectively 0.314 and 0.302. On AP(0.3), AlphaPose is the best model with 0.785. With this looser metric, human pose estimators perform better. Indeed, in the OR environment, when clinicians wear mask and hats, face detectors cannot rely on the same features as in the outside environment, such as the mouth shape and the nose. Human pose estimators, which also detect other body keypoints, are more robust than face detectors. However, they do not localize the bounding boxes accurately enough to perform well on stricter metrics. For comparison, we also provide the results on the original MVOR dataset, where the eyes are blurred, in Table \ref{mvor_public}. The results show a significant drop in the performance highlighting the importance of the eyes for face detection. \\
Overall, results of state-of-the-art detectors show a large margin for improvement on the MVOR-Faces dataset. With an IoU of 0.5, which is a less strict metric, the AP of the best model is only 0.556. On the WIDER Faces dataset, tiny face detector \cite{Hu_2017_CVPR} achieves 0.819 using the same metric, while SSH \cite{najibi2017ssh} reaches 0.944 and S3FD \cite{1708.05237} 0.958. Qualitative results shown in Fig. \ref{qualitative_soa} illustrate some of the mistakes made by the state-of-the-art detectors on the MVOR-Faces dataset, e.g. multiple detections, false positives, false negatives.
\subsection{Iterative Self-supervision} 
As mentioned in section 2.2, we use SSH \cite{najibi2017ssh} for the self-supervised process. We conduct several experiments on self-supervision to demonstrate the interest of this iterative approach. During training, we use the following hyper-parameters: stochastic gradient descent with a learning rate of 0.04, momentum of 0.9 and weight decay of $5e^{-4}$. The batch size is 2. Anchors, which correspond to a location (x,y) in the image and a predefined bounding box size (width, height), are considered as positives if their IoU with a ground-truth bounding box is greater than 0.5, as negatives otherwise. During inference, we use an image pyramid of four levels, as the authors. The aspect ratio of each rescaled image is preserved. The weights are initialized with the ones provided by the authors, after training on the WIDER Faces dataset.\\
In Table \ref{iterative_process}, we provide the test-results of our proposed iterative process, with different hyper-parameters (number of iterations and number of training batches used before relabelling). When relabelling, we filter the detections with the same criteria as explained in section 2.2, i.e. 2*N best detections where N is the number of images. The training is done with the same parameters as mentioned above. The model which performs best on the test dataset is achieved at iteration 4 when training with 2000 batches before regenerating the labels. The model outperforms the state-of-the-art with a  large margin on all metrics. On AP(0.5), it outperforms tiny face detector \cite{Hu_2017_CVPR} by more than 9\%, and the original SSH model by 13.1\%. \\ 
In Fig. \ref{Qualitative results}, we compare a few detections from the original SSH and the best self-supervised model on the test-set. The latter detects much harder examples, with occlusion or uncommon poses, and has fewer false positives. \\
In Table \ref{baseline}, we show an ablation study of self-supervision for domain adaptation, with no relabelling of target images by the self-supervised model. The 20k images of the unlabeled dataset are annotated with the detections of the original SSH model. We filter the predictions with the same criteria: since we have 20k images, we take the best 40k detections. Then, the model is fine-tuned by training on synthetically annotated images on 15k training batches. We observe a quick saturation on the test-set, MVOR-Faces: the AP(0.3:0.95) reaches 0.372 after 1k batches and 0.378 after 10k batches (i.e., with one epoch on the entire unlabeled dataset). At 15k batches, the AP(0.3:0.95) is back at 0.372. The quick saturation of this process highlights the interest of our iterative approach. \\
\begin{table}[]
\centering
\begin{tabular}{|c|cccc|}
\hline
Detector    & AP(0.3:0.95) & AP(0.3) & AP(0.5) & AR(0.3:0.95) \\ \hline
Faster-RCNN \cite{ren2015faster,1606.03473} & 0.254        & 0.651   & 0.407   & 0.345        \\ 
S3FD  \cite{1708.05237}      & 0.302        & 0.627   & 0.486   & 0.395        \\
Tiny Face \cite{Hu_2017_CVPR}   & 0.340 &	0.734 &	0.556 &	0.428       \\
SSH \cite{najibi2017ssh}        & 0.314        & 0.704   & 0.517   & 0.421        \\ \hline
AlphaPose \cite{fang2017rmpe} & 0.279        & 0.785   & 0.463   & 0.358        \\
OpenPose \cite{cao2016realtime}   & 0.240        & 0.776   & 0.365   & 0.316 \\ \hline
Self-supervised SSH & \textbf{0.402} &  \textbf{0.800} &  \textbf{0.648} &  \textbf{0.474} \\\hline
\end{tabular}
\caption{Results of state-of-the-art face detectors on MVOR-Faces. First four methods are bounding box based face detectors. AlphaPose and OpenPose are human pose estimators, from which face bounding boxes are generated from the face keypoints. Results show the margin for improvement on the MVOR-Faces dataset.}
\label{state_of_the_art}
\end{table}

\begin{table}[]
\centering
\begin{tabular}{|c|cccc|}
\hline
Detector    & AP(0.3:0.95) & AP(0.3) & AP(0.5) & AR(0.3:0.95) \\ \hline
Tiny Face \cite{Hu_2017_CVPR}  & 0.237 & 0.627 & 0.369 & 0.331       \\
SSH   \cite{najibi2017ssh}      & 0.229 & 0.600 & 0.368 & 0.368        \\ \hline
AlphaPose \cite{fang2017rmpe}  & 0.239 & \textbf{0.742} &  0.370 & 0.323 \\ \hline
Self-supervised SSH & \textbf{0.306} & 0.711 &  \textbf{0.492} &  \textbf{0.403} \\ \hline
\end{tabular}
\caption{Comparative study: results of state-of-the-art face detectors on MVOR \cite{srivastav2018mvor}, the public version of MVOR-Faces. Here, clinicians wearing a mask are blurred around the eyes. Results show the significant decrease in the performance as compared to MVOR-Faces shown in Table \ref{state_of_the_art}.}
\label{mvor_public}
\end{table}

\begin{table}[]
\centering
\begin{tabular}{|c|cccc|}
\hline
Number of training batches & AP(0.3:0.95) & AP(0.3) & AP(0.5) & AR(0.3:0.95) \\ \hline
Original SSH model \cite{najibi2017ssh} & 0.314        & 0.704   & 0.517   & 0.421        \\ \hline
1000 & 0.372 &  \textbf{0.781} &  0.597 & \textbf{0.462} \\
2000 & 0.373 &  0.769 & 0.593 & 0.458 \\
3000 & 0.374 &  0.770 & 0.595 & 0.458 \\
5000 & 0.372 &  0.770 & 0.598 & 0.454\\
10000 & \textbf{0.378} &  0.778 & \textbf{0.608} &  0.462 \\
15000 & 0.372 & 0.781 & 0.600 & 0.457 \\\hline
\end{tabular}
\caption{Comparative study: training SSH model without re-generating the synthetic labels. One training batch is composed of two images. Here, the self-supervised model is trained on the images annotated by the original SSH model and initialized with SSH weights. The AP saturates quite fast. After 1k batches, the AP does not significantly increase. The best results are much lower than best results with the iterative approach in Table \ref{fig:iterative_approach}. }
\label{baseline}
\end{table}

\begin{table}[]
\centering
\begin{tabular}{|c|p{2cm}|cccc|}
\hline
Iteration & Number of training batches before relabelling & AP(0.3:0.95) & AP(0.3) & AP(0.5) & AR(0.3:0.95) \\ \hline
& Original SSH model \cite{najibi2017ssh} & 0.314        & 0.704   & 0.517   & 0.421        \\ \hline
1 & \multirow{4}{*}{1000} & 0.373 & 0.782 & 0.598 & 0.462 \\
2 &  & 0.365 &  0.783 & 0.590 & 0.461                  \\
3 &  &  0.367 & 0.785 & 0.592 & 0.450                 \\ 
4 & & 0.365 & 0.788 & 0.594 & 0.456 \\ \hline
1 & \multirow{4}{*}{2000}   & 0.373 & 0.773 & 0.596 & 0.459 \\ 
2 & & 0.385 & 0.796 & 0.622 & 0.466 \\
3 & & 0.393 & \textbf{0.808} & 0.630 & 0.465 \\
4 & & \textbf{0.402} & 0.800 & \textbf{0.648} & \textbf{0.474} \\ \hline
1 & \multirow{4}{*}{3000} & 0.373 & 0.772 & 0.595 & 0.457 \\
2 &  & 0.377 &  0.793 & 0.608 & 0.452 \\
3 &  & 0.383 &  0.806 & 0.614 & 0.459  \\
4 & &  0.371 &  0.797 & 0.604 & 0.448 \\ \hline
\end{tabular}
\caption{The iterative process of self-supervision with different hyper-parameters. One iteration consists of three steps: (1) Generate predictions on the unlabeled dataset with the last model. (2) Filter detections: select the best 2N detections on N images. (3) Retrain the model on 1k, 2k or 3k training batches. When training with 2k or 3k batches before relabelling, it improves the results from the baseline approach (see in Table \ref{baseline}). One batch is composed of two images. }
\label{iterative_process}
\end{table}


\section{Conclusion}
We propose the first broad evaluation of state-of-the-art face detectors on OR images. Since the results show a large margin for improvement, we also propose to use an iterative self-supervised approach to adapt a face detector to a given OR. It consists of gathering images of the target environment, generating synthetic annotations with a model trained on a manually annotated dataset, and retraining it iteratively using the synthetic labels. This method is generic and applicable to any OR configuration. Our self-supervised detector outperforms the state-of-the-art on MVOR-Faces by a large margin, namely by more than 6\% on AP(0.3:0.95). By significantly improving the accuracy of face detection, we show that self-supervision is a promising direction to transfer state-of-the-art computer vision approaches to the medical context, where annotations are challenging to generate.

\begin{acknowledgements}
This work was supported by French state funds managed by the ANR within the Investissements d'Avenir program under references ANR-16-CE33-0009 (DeepSurg), ANR-11-LABX-0004 (Labex CAMI) and ANR-10-IDEX-0002-02 (IdEx Unistra). The authors would also like to thank the members of the Interventional Radiology Department at University Hospital of Strasbourg for their help in generating the dataset.
\end{acknowledgements}


\bibliographystyle{unsrt}
\bibliography{references}
\end{document}